\documentclass[letterpaper, 10 pt, conference]{ieeeconf}
\IEEEoverridecommandlockouts    % This command is only needed if
\usepackage{fancyhdr}
\fancypagestyle{withfooter}{
  
  \fancyfoot[C]{\footnotesize Accepted as Extended Abstract to the IEEE IROS 2024 Workshop on AI and Robotics For Future Farming}
}
\usepackage{graphics}           
\usepackage{times}              
\usepackage{amsmath}            
\usepackage{amssymb}            
\usepackage{graphicx}
\usepackage{algorithm}
\usepackage[noend]{algpseudocode}
\usepackage{booktabs}
\usepackage{color}
\usepackage{listings}
\usepackage{subfiles}
\usepackage{hyperref}
\usepackage[none]{hyphenat}
\usepackage[nocompress]{cite} % sorts citations automatically, e.g., "[1],[2],[3]" instead of "[2],[3],[1]".
\definecolor{instructioncolor}{rgb}{.5,.5,.5}

\usepackage[font=small]{caption}

\def\eqref#1{Eq.~(\ref{#1})}

\makeatletter
\usepackage{xspace}
\DeclareRobustCommand\onedot{\futurelet\@let@token\@onedot}
\def\@onedot{\ifx\@let@token.\else.\null\fi\xspace}

\makeatother

\usepackage{array}
\newcolumntype{L}[1]{>{\raggedright\let\newline\\\arraybackslash\hspace{0pt}}m{#1}}
\newcolumntype{C}[1]{>{\centering\let\newline\\\arraybackslash\hspace{0pt}}m{#1}}
\newcolumntype{R}[1]{>{\raggedleft\let\newline\\\arraybackslash\hspace{0pt}}m{#1}}

\title{\LARGE \bf AdaCropFollow: Self-Supervised Online Adaptation\\for Visual Under-Canopy Navigation}

\author{Arun N. Sivakumar$^{1}$, Federico Magistri$^{2}$, Mateus V. Gasparino$^{1}$, Jens Behley$^{2}$, \\ Cyrill Stachniss$^{2,3,4}$, Girish Chowdhary$^{1}$% <-this % stops a space
\thanks{$^{1}$Field Robotics Engineering and Sciences Hub (FRESH), University of Illinois Urbana-Champaign (UIUC), IL}%
\thanks{$^{2}$Center for Robotics, University of Bonn, Germany}%
\thanks{$^{3}$Department of Engineering Science at the University of Oxford, UK}%
\thanks{$^{4}$Lamarr Institute for Machine Learning and Artificial Intelligence, Germany}%
\thanks{This work was supported in part by AIFARMS \#1024178 ,NSF-USDA COALESCE \#2021-67021-34418, NSF NRI 2.0 NIFA \#2021-67021-33449, USDA grants iCOVER(\#NR233A750004G066), iFARM(\#2022-77038-37306), and NSF STTR \#1951250.}
}

\begin{document}

\maketitle
\thispagestyle{withfooter}
\pagestyle{withfooter}
% \thispagestyle{empty}
% \pagestyle{empty}

%%%%%%%%%%%%%%%%%%%%%%%%%%%%%%%%%%%%%%%%%%%%%%%%%%%%%%%%%%%%%%%%%%%%%%%%%%%%%%%%
\begin{abstract}
Under-canopy agricultural robots can enable various applications like precise monitoring, spraying, weeding, and plant manipulation tasks throughout the growing season. Autonomous navigation under the canopy is challenging due to the degradation in accuracy of RTK-GPS and the large variability in the visual appearance of the scene over time. In prior work, we developed a supervised learning-based perception system with semantic keypoint representation and deployed this in various field conditions. A large number of failures of this system can be attributed to the inability of the perception model to adapt to the domain shift encountered during deployment. In this paper, we propose a self-supervised online adaptation method for adapting the semantic keypoint representation using a visual foundational model, geometric prior, and pseudo labeling. Our preliminary experiments show that with minimal data and fine-tuning of parameters, the keypoint prediction model trained with labels on the source domain can be adapted in a self-supervised manner to various challenging target domains onboard the robot computer using our method. This can enable fully autonomous row-following capability in under-canopy robots across fields and crops without requiring human intervention. 

\end{abstract}

%%%%%%%%%%%%%%%%%%%%%%%%%%%%%%%%%%%%%%%%%%%%%%%%%%%%%%%%%%%%%%%%%%%%%%%%%%%%%%%%
\section{Introduction}
\label{sec:intro}

Agriculture faces major challenges due to the increasing demand for food, fiber, and energy from the growing world population, the shrinking availability of farmland and labor, and the environmental concerns due to over application of farm inputs. Autonomous robots can help tackle these challenges by sustainably improving productivity in agricultural production as well as in crop improvement research. In particular, compact and low-cost under-canopy agricultural robots can enable various applications such as precise monitoring, high throughput phenotyping, spraying, weeding, and plant manipulation throughout the growing season.

Under-canopy navigation is challenging due to factors such as degradation in RTK-GPS accuracy under the canopy, significant occlusion and clutter in this environment, and less margin for error due to the minimal traversable space available for the robot. In prior work \cite{Sivakumar-RSS-24}, we developed a vision and learning-based modular navigation system with semantic keypoint representation and deployed this system over large distances with significantly better performance in terms of the number of collisions than the prior state of the art. However, the keypoint prediction model was trained with supervised learning using a large labeled dataset of crop row lines and used as a frozen model with no continuous adaptation during deployment. A large number of failures observed during the deployment were attributed to the errors in keypoint prediction stemming from the inability to adapt to novel scenarios in a self-supervised way.

In this paper, we investigate the problem of self-supervised online adaptation of the keypoint model for under-canopy navigation. This implies that by using minimal data from a target domain, we update the keypoint model with the compute available onboard the robot in a self-supervised manner. The main research challenges associated with this problem include identifying an appropriate self-supervised loss function for the task, finding the minimal number of parameters that can be updated to adapt the network while considering the compute constraints of the robot and preventing catastrophic forgetting of prior knowledge which is a common issue when fine-tuning networks with limited data.

\begin{figure}[t]
  \centering
  \includegraphics[width=1\linewidth]{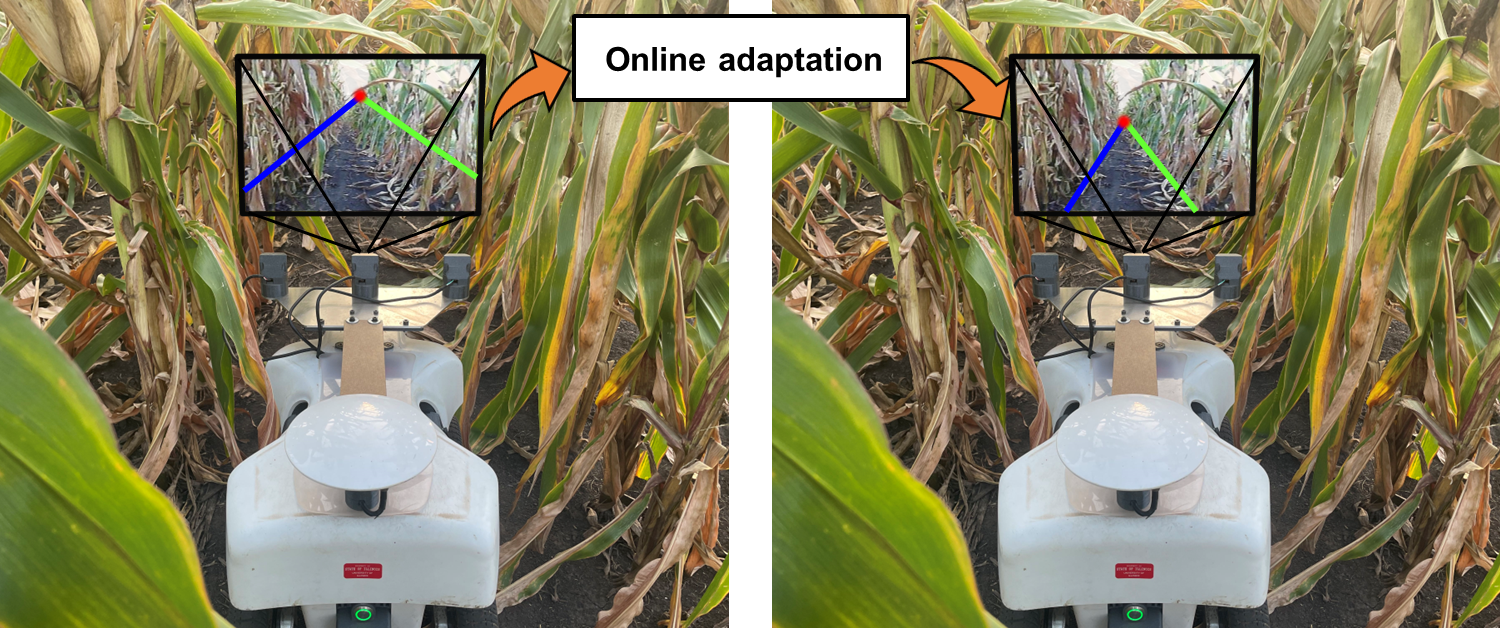}
  \caption{Under-canopy robots navigating with machine learning-based systems encounter domain shifts due to variations in this environment. We propose a self-supervised online learning method to adapt to this.}  \label{fig:motivation}
\end{figure}

% \textbf{How \& What:}

\begin{figure*}[t]
  \centering
  \includegraphics[width=0.95\linewidth]{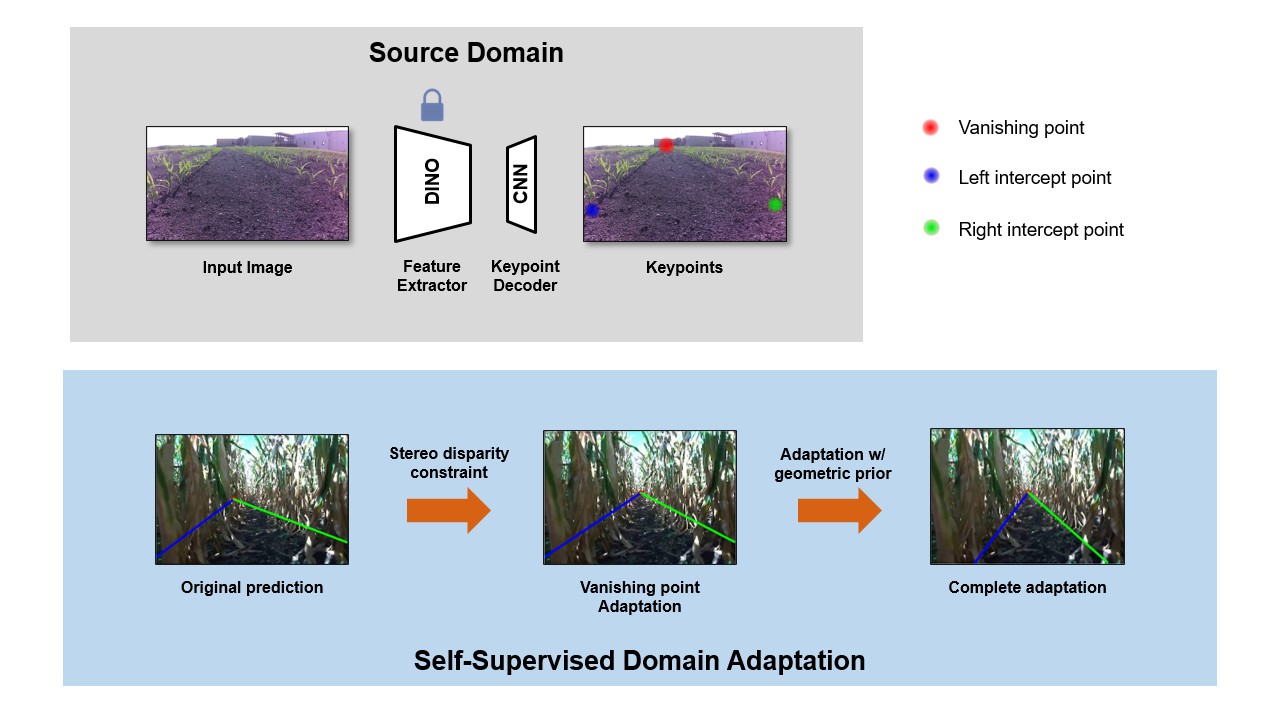}
  \caption{AdaCropFollow overview: We train the keypoint prediction model in the source domain (early season corn) with labeled data. During online adaptation, we first adapt the vanishing point using stereo disparity loss and then adapt the remaining two intercept keypoints using pseudo-labeling guided by geometric prior. Note that we use the frozen feature encoder from Dinov2 during the source domain as well as online adaptation.}
  \label{fig:overview}
\end{figure*}

A common approach for self-supervised adaptation is to use entropy minimization based on pseudo-labeling \cite{lee2013pseudo,sohn2020fixmatch,wang2021tent}. However, the approach of choosing the pseudo-labels just based on the entropy of the predictions results in noisy labels when the domain shift is significant. We assume that we have access to stereo images during adaptation and also that the information on camera intrinsics and extrinsics, as well as an estimate of the distance between the crop rows, is known. With this information, we define a geometric prior that is used as the criterion to select the pseudo-labels for model adaptation. Also, we use the visual foundational model Dinov2 \cite{oquab2023dinov2} as our frozen feature encoder similar to Frey et al. \cite{frey23fast} which enables updating only the minimal number of parameters in the convolutional layers of the keypoint decoder to adapt the network. This also helps overcome the problem of catastrophic forgetting.

The main contribution of this paper is a novel method for self-supervised online adaptation of an under-canopy keypoint prediction model with minimal data and compute by using the encoder from a visual foundational model and geometric prior guided pseudo-labeling.

%%%%%%%%%%%%%%%%%%%%%%%%%%%%%%%%%%%%%%%%%%%%%%%%%%%%%%%%%%%%%%%%%%%%%%%%%%%%%%%%
\section{Our Approach}
\label{sec:main}
We use the same semantic keypoint representation for perception as in our prior work \cite{Sivakumar-RSS-24}. The traversable area for the robot is represented by the triangle formed on the image by the two vanishing lines representing the crop rows in 3D. We denote the vanishing point formed by the intersection of the vanishing lines and two intercept points formed by the intersection of each crop line with the bottom half of the image as our semantic keypoints of interest (shown in Fig \ref{fig:overview}).  We assume that we have access to a model pre-trained for this task on a source domain with labeled data. Our method for online adaptation consists of two stages. We first adapt the vanishing point in the stage followed by the intercept points in the second stage as detailed in the following subsections.

\subsection{Adapting vanishing point}
During the online adaptation in target domain, we assume access to stereo images. We adapt the vanishing point in the first stage. Since the vanishing point is a point at infinity in the 3D space, we know that the pixel coordinate of its projection in both left and right images of a rectified strereo pair should be the same which implies that the disparity should be equal to zero. Therefore we use the disparity in the pixel location of the vanishing point in the left and right image as the loss for adapting the vanishing point.

\subsection{Geometric prior}
In the next stage, we first define the geometric prior from the known information of camera intrinsics and extrinsics, the distance between the crop rows, the width of the robot, and the height of the camera from the ground. Given these, for the case of pitch, roll, and heading of the robot being zero, we can calculate the following three constraints using projective geometry i) The upper and lower bound for the left and right intercept keypoints on both the images (corresponding to the cases of the robot being close to the left crop row and right crop row respectively) ii) The width of the base of the triangle representing the traversable area iii) The disparity of the intercept keypoints between the left and right image. These three constraints together constitute our geometric prior. Note that these constraints can only be defined for the canonical triangle representing the case of the robot roll, pitch, and heading being zero. 

\begin{figure*}[t]
  \centering
  \includegraphics[width=0.95\linewidth]{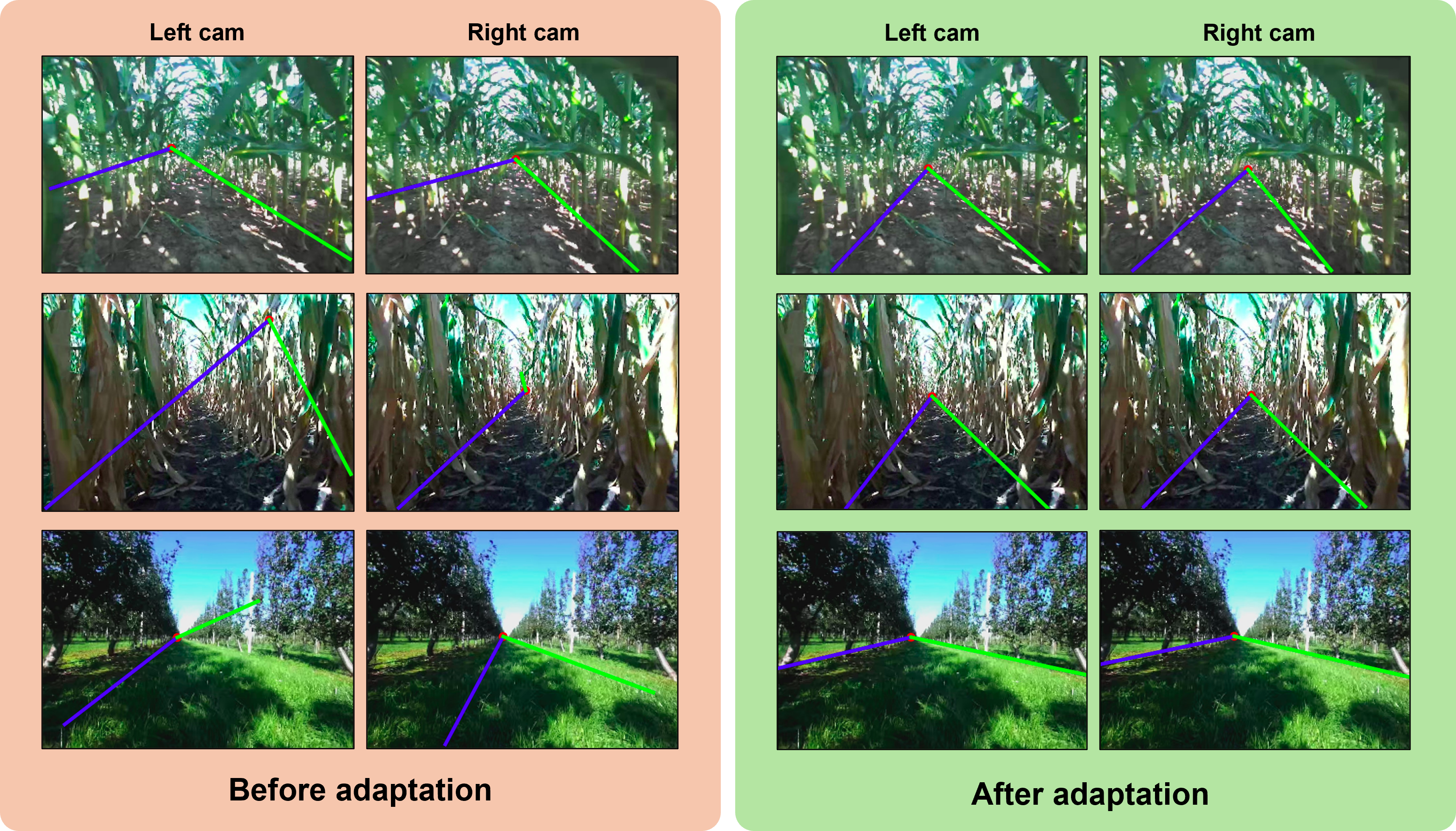}
  \caption{We visualize the pixel locations predicted by the model before and after adaptation in three different target environments namely green late-season corn, brown late-season corn, and orchard environment. We can see considerable improvement in accuracy of predicted pixel location of keypoints after adaptation.}
  \label{fig:results_vis}
\end{figure*}
\subsection{Generating intercept keypoints pseudo-labels and fine-tuning the network}
We assume access to the data of roll angle of the robot from IMU. After the vanishing point adaptation stage, the vanishing point predictions are assumed to be good. We start by running inference of the vanishing point adapted network on the target domain data to get the predictions for all the keypoints. We calculate the pitch of the robot by rotating the vanishing point to correct for roll. We then calculate the heading of the robot by applying a homography to correct for the pitch of the robot if non-zero. Now given that we have access to roll, pitch, and heading of the robot, we construct a homography matrix to represent this rotation. This homography matrix is applied on the intercept keypoint predictions for all data points to transform them into the convention of the canonical triangle. We then use the three constraints defined in the geometric prior to identify the good predictions that can be treated as pseudo-labels. If the predictions of one of the stereo images satisfy the constraints for pseudo-labeling, we use the geometric prior to get the corresponding labels for the other image as well. We finetune the keypoint decoder network with the identified pseudo-labels.

%%%%%%%%%%%%%%%%%%%%%%%%%%%%%%%%%%%%%%%%%%%%%%%%%%%%%%%%%%%%%%%%%%%%%%%%%%%%%%%%
\section{Experiments and Results}
\label{sec:exp}

The main focus of this work is on self-supervised online adaptation of the under-canopy keypoint perception model to various target domains. In the following subsections, we first explain the details of our network architecture and data augmentation during supervised training in the source domain and self-supervised online adaptation in the target domain. We then present our preliminary results as qualitative visualizations of model predictions before and after adaptation in three different target domains. We also present a quantitative comparison with mean $L_1$ error in the predicted pixel location of keypoints as the metric in brown late-season corn as the target domain for adaptation. 

\subsection{Supervised training in source domain}
We choose the relatively simple environment of early-season corn fields as our source domain. We use frozen Dinov2 as our feature encoder \cite{oquab2023dinov2} and train two convolutional upsampling blocks followed by a head layer to predict the three semantic keypoints as spatial heatmaps. The convolutional upsampling blocks also contain batch norm layers in them. We only use data from monocular cameras with flip augmentation during this phase. The two covolutional upsampling layers contain 590336 and 110848 parameters respectively followed by 195 parameters in the head layer. We only update these parameters during this source domain training.

\subsection{Self-supervised online training in target domain}
We perform the self-supervised online training in two stages. We adapt the vanishing point in the first stage. We only perform this adaptation for the vanishing point if the disparity in pixel location of the vanishing between the left and right images is greater than 10 for more than 50\% of images in the target domain dataset. This prevents the network from converging to undesired local minima in target domains in which the source domain-trained model already generalizes well for vanishing points. In addition, we also use a large batch size of 64 and a weight decay penalty of 0.01 to smooth the gradient and prevent sudden changes in network weights. During the second stage, we adapt the intercept keypoints. After generating the pseudo-labels using the constraints defined by the geometric prior, we finetune all the parameters of the head layer and the second upsampling block as well as the parameters of the batch norm layer in the first upsampling block. In total 111360 parameters are updated during this pseudo-labeling-based finetuning phase. We use flip data augmentation during this fine-tuning and we repeat this pseudo-labeling process for five iterations during which the number of images that satisfy the criterion improves significantly. 

\subsection{Qualitative evaluation}
For qualitative evaluation, we show in Fig \ref{fig:results_vis} the location of the three keypoints predicted by the model before and after adaptation in three significantly different target domains such as green late-season corn, brown late-season corn, and orchard environment. On visual inspection, we can see that the accuracy of predicted pixel locations improves considerably after adaptation in all three environments.

\subsection{Quantitative evaluation}
We also quantitatively evaluated the keypoint prediction accuracy before and after adaptation in brown color late-season corn by using mean $L_1$ error as the metric. We labeled a small subset of $50$ images not used for adaptation and used it for validation. Table \ref{tab:l1_error} shows these results. The mean $L_1$ error reduces considerably after adaptation to target domain using our method.

\begin{table}[t]
  \caption{Mean $L_1$ error for the semantic keypoints before and after adaptation}
  \centering

  \begin{tabular}{C{3cm}C{2cm}C{2cm}}
    \toprule
    \textbf{Keypoint} & \textbf{Before adaptation Mean $L_1$ error}                                  & \textbf{After adaptation Mean $L_1$ error}                                 \\

    \midrule
    Vanishing point  & 25.04 & \textbf{9.18} \\
    Left intercept point  & 92.58 & \textbf{14.78} \\
    Right intercept point  & 164.34 & \textbf{23.34} \\
    \bottomrule
  \end{tabular}

  \label{tab:l1_error}
\end{table}

%%%%%%%%%%%%%%%%%%%%%%%%%%%%%%%%%%%%%%%%%%%%%%%%%%%%%%%%%%%%%%%%%%%%%%%%%%%%%%%%
\section{Conclusion}
\label{sec:conclusion}

In this paper, we presented a novel approach to self-supervised online adaptation for under-canopy navigation. Our method exploits the frozen encoder from a visual foundation model and the known geometric prior to adapt to different target domains with minimal data. Our preliminary offline evaluation results in terms of qualitative visualization of predicted pixel locations as well as quantitate evaluation in mean $L_1$ error in pixel location show considerable improvement in predictions after adaptation. These results show promise for field deployment of our method.

\bibliographystyle{plain_abbrv}

\bibliography{new}

\end{document}